\def\paperID{8354}
\def\confName{CVPR}
\def\confYear{2025}

\def\paperTitle{DEFOM-Stereo: Depth Foundation Model Based Stereo Matching}

\def\authorBlock{
    Hualie Jiang$^{1}$ \quad
    Zhiqiang Lou$^{1}$ \quad
    Laiyan Ding$^{2}$ \\
    Rui Xu$^{1}$\thanks{corresponding author} \quad
    Minglang Tan$^{1}$ \quad
    Wenjie Jiang$^{1}$ \quad
    Rui Huang$^{2}$ \\
    $^{1}$Insta360 Research \qquad $^{2}$The Chinese University of Hong Kong, Shenzhen\\
   {\tt\small \{jianghualie, johnlou, jerry1\}@insta360.com, laiyanding@link.cuhk.edu.cn}\\
   \small\url{https://insta360-research-team.github.io/DEFOM-Stereo}
}

\newif\ifreview 
\newif\ifarxiv \newcommand{\arxiv}{\arxivtrue}
\newif\ifcamera 
\newif\ifrebuttal 

\arxiv 

\pdfoutput=1
\documentclass[10pt,twocolumn,letterpaper]{article}
\ifreview \usepackage[review]{cvpr} \fi
\ifarxiv \usepackage[pagenumbers]{cvpr} \fi
\ifrebuttal \usepackage[rebuttal]{cvpr} \fi
\ifcamera \usepackage{cvpr} \fi

\usepackage{graphicx}	
\usepackage{amsmath}	
\usepackage{amssymb}	
\usepackage{booktabs}
\usepackage{times}
\usepackage{microtype}
\usepackage{epsfig}
\usepackage[table,xcdraw,dvipsnames]{xcolor}
\usepackage{caption}
\usepackage{float}
\usepackage{placeins}
\usepackage{color, colortbl}
\usepackage{stfloats}
\usepackage{enumitem}
\usepackage{tabularx}
\usepackage{xstring}
\usepackage{multirow}
\usepackage{xspace}
\usepackage{url}
\usepackage{subcaption}
\usepackage{xcolor}
\usepackage[hang,flushmargin]{footmisc}
\usepackage{overpic}

\ifcamera \usepackage[accsupp]{axessibility} \fi

% ...

\ifarxiv  \fi

\newcommand{\R}[1]{{%
    \textbf{%
        \ifstrequal{#1}{1}{\textcolor{red}{R#1}}{%
        \ifstrequal{#1}{2}{\textcolor{blue}{R#1}}{%
        \ifstrequal{#1}{3}{\textcolor{magenta}{R#1}}{%
        \ifstrequal{#1}{4}{\textcolor{teal}{R#1}}{%
                           \textcolor{cyan}{R#1}%
        }}}}%
    }%
}}

\usepackage{xr-hyper}

\makeatletter
\newcommand*{\addFileDependency}[1]{
  \typeout{(#1)}
  \@addtofilelist{#1}
  \IfFileExists{#1}{}{\typeout{No file #1.}}
}

\makeatother
\newcommand*{\myexternaldocument}[1]{
    \externaldocument{#1}
    \addFileDependency{#1.tex}
    \addFileDependency{#1.aux}
}

\definecolor{cvprblue}{rgb}{0.21,0.49,0.74}
\usepackage[pagebackref,breaklinks,colorlinks,citecolor=cvprblue]{hyperref}
\usepackage[capitalize]{cleveref}
\crefname{section}{Sec.}{Secs.}
\crefname{table}{Table}{Tables}
\crefname{figure}{Fig.}{Figs.}

\ifarxiv \crefname{appendix}{App.}{Apps.}
\else \crefname{appendix}{Suppl.}{Suppls.} \fi

\unless\ifarxiv \myexternaldocument{_supplementary} \fi

\begin{document}
\title{\paperTitle}
\author{\authorBlock}
\maketitle

\begin{abstract}
Stereo matching is a key technique for metric depth estimation in computer vision and robotics. 
Real-world challenges like occlusion and non-texture hinder accurate disparity estimation from binocular matching cues.
Recently, monocular relative depth estimation has shown remarkable generalization using vision foundation models.
Thus, to facilitate robust stereo matching with monocular depth cues, we incorporate a robust monocular relative depth model into the recurrent stereo-matching framework, building a new framework for depth foundation model-based stereo-matching, DEFOM-Stereo. 
In the feature extraction stage, we construct the combined context and matching feature encoder by integrating features from conventional CNNs and DEFOM. In the update stage, we use the depth predicted by DEFOM to initialize the recurrent disparity and introduce a scale update module to refine the disparity at the correct scale.
 DEFOM-Stereo is verified to have much stronger zero-shot generalization compared with SOTA methods. Moreover, DEFOM-Stereo achieves top performance on the KITTI 2012, KITTI 2015, Middlebury, and ETH3D benchmarks, ranking $1^{st}$ on many metrics. In the joint evaluation under the robust vision challenge, our model simultaneously outperforms previous models on the individual benchmarks, further demonstrating its outstanding capabilities.
 \end{abstract}

\section{Introduction}
\label{sec:intro}

\begin{figure}[t]
    \centering{
    \input{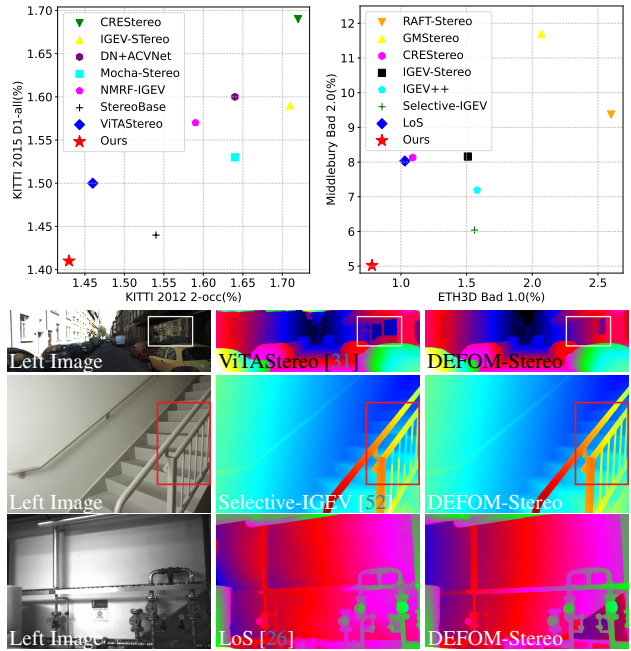}}
    \vspace{-15pt}
    \caption{\textbf{Row 1:} Comparisons with SOTA stereo methods on KITTI 2012~\cite{geiger2012we} and 2015~\cite{menze2015object}, Middlebury~\cite{scharstein2014high} and  ETH3D~\cite{schops2017multi} leaderboards. \textbf{Row 2-4:} Visual comparison with ViTAStereo~\cite{liu2024playing} on KITTI 2012, Selective-IGEV~\cite{wang2024selective} on Middlebury, and LoS~\cite{li2024local} on ETH3D. Our method performs robustly on reflective, textureless, and ill-exposure areas and recovers more structural details.}
    \label{fig:ranking}
\vspace{-10pt}
\end{figure}

Stereo matching involves recovering the disparity between two views and has been a fundamental topic in computer vision for decades.
Stereo matching has obtained steady improvement with the consistently proposed algorithms, which range from hand-designed ones~\cite{scharstein2002taxonomy} to end-to-end deep architectures~\cite{laga2020survey, tosi2024survey}. The deep matching frameworks have mainly shifted from cost-volume filtering ones~\cite{gcnet, psmnet, gwcnet, xu2020aanet, zhang2020domain} to recent recurrent refinement ones~\cite{lipson2021raft, li2022practical, zhao2023high, xu2023iterative, chen2024mocha, wang2024selective}. According to~\cite{tosi2024survey}, despite RAFT-Stereo~\cite{lipson2021raft} has made good progress, the generalization and robustness remain challenging. The critical challenges lie in occlusion, non-texture, image blur, high-resolution, \etc. \textit{etc}.

On the other hand, a closely related task, monocular relative depth estimation~\cite{ranftl2020towards, ranftl2021vision, birkl2023midas, ke2024repurposing, depthanything, depth_anything_v2}~has achieved advancing zero-shot generalization ability across diverse scenes. Among them, Depth Anything V2~\cite{depth_anything_v2}, is a model with a pre-trained vision transformer (ViT)~\cite{dosovitskiy2020image} backbone of the general-purpose vision foundation model DINOv2~\cite{oquab2023dinov2} and a head of the dense prediction transformer (DPT)~\cite{ranftl2021vision}. It is trained on synthetic labeled images and distilled on large realistic unlabeled images to conduct robust and efficient depth prediction and recover impressive fine-grained details. These properties allow it to be a depth foundation model (DEFOM). Nevertheless, DEFOM does not provide metric depth, which is crucial for robotic applications. 
To achieve more robust stereo depth estimation, we incorporate Depth Anything V2 into RAFT-Stereo~\cite{lipson2021raft} to build a DEFOM-based stereo matching (DEFOM-Stereo) model. To be specific, we perform the incorporation with two aspects, 1) utilizing DEFOM's pre-trained features to improve the feature extraction for stereo matching, and 2) facilitating disparity updating with the predicted monocular depth.

Utilizing the pre-trained ViT to improve stereo matching has recently been explored in \cite{liu2024playing, zhang2024exploring} via developing attention-based adapters to convert the ViT backbone features into matching features. In contrast, the DEFOM provides dense feature maps with a DPT head, and the feature is more relevant to the depth task.  We demonstrate that a simple convolutional fusion of DPT features and plain CNN features is effective. As the existing DPT should be fixed to predict depth, we initialize a trainable DPT head to provide more flexible features for fusion. Apart from the matching feature, there is a context encoder of the recurrent stereo framework to provide monocular cues controlling the recurrent disparity update process. Thus, we propose a combined matching feature and context encoder with CNN and DEFOM. 

\begin{figure}[t]
    \centering{
    \begin{subfigure}{0.32\linewidth}
        \centering
        \begin{overpic}[width=1.0\linewidth]{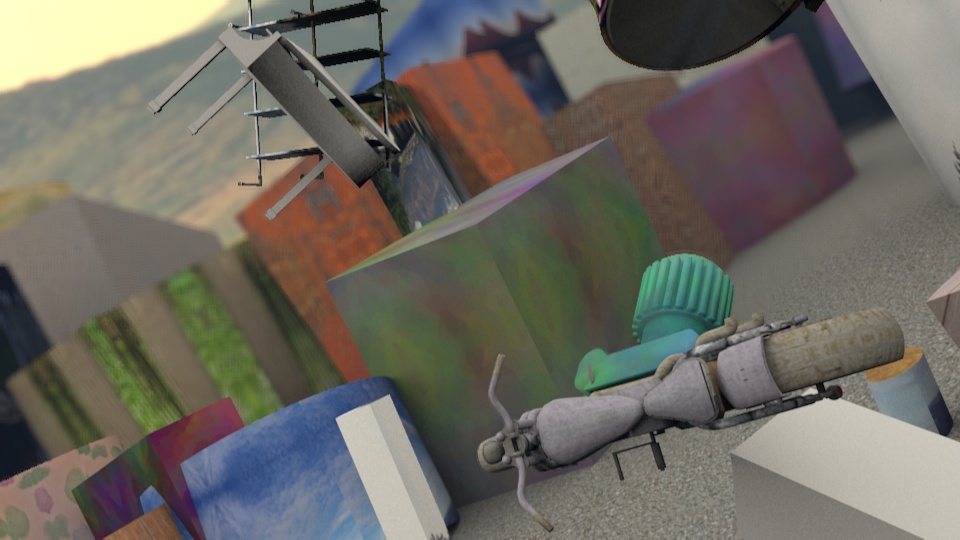}
            \put(1, 2){\textcolor{white}{\footnotesize{Left Image}}}
        \end{overpic}
    \end{subfigure}
    \begin{subfigure}{0.32\linewidth}
        \centering
        \begin{overpic}[width=1.0\linewidth]{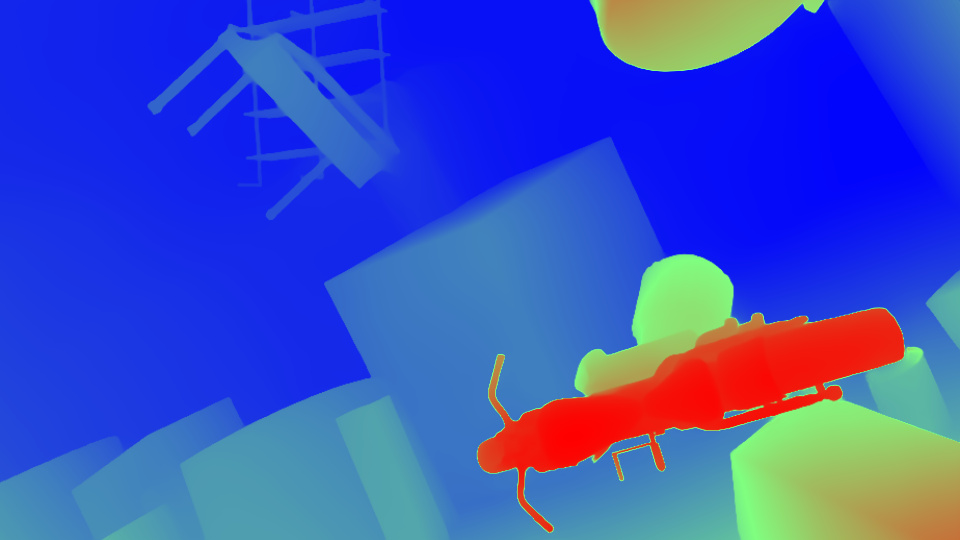}
            \put(1, 2){\textcolor{white}{\footnotesize{Depth Anything V2}}}
        \end{overpic}
    \end{subfigure}
    \begin{subfigure}{0.32\linewidth}
        \centering
        \begin{overpic}[width=1.0\linewidth]{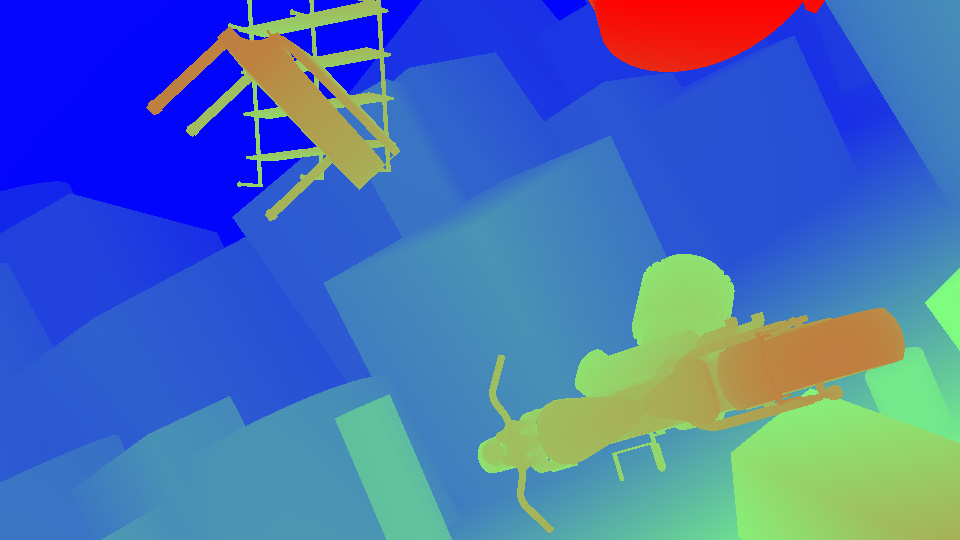}
            \put(0, 2){\textcolor{white}{\footnotesize{Ground-Truth Disparity}}}
        \end{overpic}
    \end{subfigure}
    }
    \vspace{-2px}
    \caption{Illustration of the significant scale inconsistency in the depth estimate of Depth Anything V2~\cite{depth_anything_v2} compared to the ground-truth disparity on the synthetic Scene Flow dataset.}
    \label{fig:scale_inconsistency}
\vspace{-15pt}
\end{figure}

Besides its features, the DEFOM depth can intuitively be used to improve disparity recovery. The DEFOM depth is an affine disparity with unknown scale and shift. However, finding a single scale and shift values does help to recover accurate disparity. A preliminary evaluation of Depth Anything V2~\cite{depth_anything_v2} on typical stereo datasets using the least-square affine alignment illustrates the disparity error is significantly large, indicating the scale inconsistency for depth prediction within the image is serious. The phenomenon is more serious for synthetic stereo datasets, for example of the synthetic Scene Flow~\cite{sceneflow} shown in Fig.~\ref{fig:scale_inconsistency}. We conjecture that the scale inconsistency is caused by the training in different FoV images with affine invariant loss. As the stereo model is usually pre-trained on the synthetic stereo dataset, the scale inconsistency of DEFOM poses challenges for recovering disparity from its depth estimate.

To cope with the scale inconsistency, we propose a scale update (SU) module that performs recurrent dense scale updates on the initialized disparity from the depth estimate. 
We found that the depth amplitude by Depth Anything V2 varies across different image samples with different backbone sizes. 
Considering that the disparity range is usually proportional to image resolution, we use the image width to normalize the depth estimate in initialization. 
To facilitate SU, we design a scale lookup (SL) on the correlation volume on which the current disparity estimate is multiplied with a set of scale values to generate sampling indexes. Compared with the conventional pyramid lookup (PL), SL has a complete search range on the whole image, enabling SU with global matching. Therefore, we insert the SU module before the conventional delta update (DU) module in RAFT-Stereo~\cite{lipson2021raft}, which can continue to recover local details upon the result of SU.

The overall framework of DEFOM-Stereo is illustrated in Fig.~\ref{fig:defom_stereo}. We have conducted extensive experiments to verify DEFOM-Stereo's effectiveness. When pre-training only on Scene Flow, DEFOM-Stereo achieves comparable performance with SOTA methods and obtains considerable progress in the zero-shot generation evaluation on KITTI 2012 and 2015, Middlebury, and ETH3D, for example, reducing the 2-pixel thresholding error rate by about a third on Middlebury. When submitting to the official benchmarks of these datasets, DEFOM-Stereo achieves top performance, ranking $1^{st}$ on many metrics on the four leaderboards among all submissions while writing this paper. A brief comparison with published SOTA methods is demonstrated in Fig.~\ref{fig:ranking}, where DEFOM-Stereo exhibits remarkable advantages. Furthermore, we also examine our model under the joint evaluation of the \href{http://robustvision.net}{Robust Vision Challenge} (RVC), and our model simultaneously outperforms all previous RVC models on the three benchmarks, particularly on KITTI 2015 and Middlebury with a large margin. 

The contributions can be summarized as follows: 
\begin{itemize}
\item We propose a novel recurrent stereo-matching framework incorporating monocular depth cues from a depth foundation model to improve robustness. 
\item We develop a simple technique that utilizes pre-trained DEFOM features to construct stronger combined feature and context encoders.
\item We present a recurrent scale update module empowered with the scale lookup, serving to recover accurate pixel-wise scales for the coarse DEFOM depth.
\end{itemize}

\section{Related Work}
\label{sec:related}

\subsection{Learning-Based Stereo Matching}
Deep learning methods have dominated stereo matching for past decades. Early methods~\cite{psmnet, gcnet, gwcnet, hsmnet, ganet, xu2020aanet} mainly construct a cost volume on defined disparities with the deep feature maps of left and right images and use 3D convolutions to perform cost aggregation. The final disparity estimate is regressed from the processed cost volume. In recent years, the recurrent refinement-based methods~\cite{lipson2021raft,li2022practical,zhao2023high,xu2023iterative,jing2023uncertainty,feng2024mc,li2024local,chen2024mocha,wang2024selective} became the mainstream, as well as the transformer-based ones~\cite{li2021revisiting, xu2023unifying, weinzaepfel2023croco}. The transformer-based methods usually use the cross-attention between two views to correlate them for the final disparity prediction. Unlike traditional cost-volume construction approaches, these methods circumvent the constraints of a pre-defined disparity range and effectively capture long-range pixel associations. More recently, a unique attempt to leverage the power of deep learning to create a neural Markov random field (NMRF-Stereo) model was presented \cite{guan2024neural}, establishing an accurate and highly efficient model.

The recurrent stereo-matching stems from the framework of recurrent all-pairs field transforms (RAFT)~\cite{teed2020raft} for optical flow estimation. RAFT-Stereo~\cite{lipson2021raft} is a pioneering work that adapts the RAFT framework for stereo matching mainly by limiting the 2D flow field to the 1D disparity field and upgrading the convolutional Gated Recurrent Unit (ConvGRU) for recurrent updates from a single level to multiple levels to enlarge the receptive field. By careful modification,  RAFT-Stereo inherits strong cross-dataset generalization from RAFT. Subsequently, there are a series of improved models from different aspects~\cite{li2022practical,zhao2023high,xu2023iterative,jing2023uncertainty,feng2024mc,li2024local,chen2024mocha,wang2024selective}. For example, CREStereo~\cite{li2022practical}~proposed a hierarchical recurrent refinement structure with an adaptive group correlation layer.  IGEV-Stereo~\cite{xu2023iterative} uses an aggregated cost volume to estimate the initial disparity and combines it with the all-pairs correlations into a combined geometry encoding volume. Selective-Stereo~\cite{wang2024selective}~preoposes a selective recurrent unit which enables the recurrent update to adaptively capture
and fuse multi-frequency information. Most of the improved models have achieved a better in-domain fitting than RAFT-Stereo. However, with comprehensive zero-shot generalization evaluation, we found that RAFT-Stereo is still more advantageous. Therefore, we use the based RAFT-Stereo framework as a baseline model and construct our model by integrating powerful monocular cues from DEFOM.

\subsection{Zero-shot Generalized Stereo Matching}
Zero-shot generalization is a crucial challenge in stereo matching, particularly for synthetic-to-real transfer, as labeled real-world datasets are difficult to obtain at scale. Earlier cost-volume-based methods~\cite{psmnet, gwcnet, ganet} exhibit a significant performance drop when trained on synthetic datasets (e.g., Scene Flow) and tested on real-world data. This decline is primarily due to the poor transferability of the matching features learned from synthetic data to real scenes. Consequently, substantial efforts have been devoted to learning domain-agnostic features. For instance, DSMNet~\cite{zhang2020domain} and FCStereo~\cite{zhang2022revisiting} aim to learn domain-invariant features. DSMNet introduces domain normalization to regularize feature distributions, while FCStereo applies a contrastive loss to ensure feature consistency between matched pixels and a stereo whitening loss to preserve feature consistency across domains. Other approaches, such as ITSA~\cite{chuah2022itsa} and HVT~\cite{chang2023domain}, tackle the issue of shortcut learning—where superficial cues are exploited instead of transferable representations—by using information-theoretic losses to constrain the encoding of shortcut-related information or by transforming synthetic training images to diversify the training domain.

In contrast to these works, which focus on strategies tailored to improve generalization, RAFT-Stereo achieves significant improvements in zero-shot performance via architectural innovation. We further enhance zero-shot generalization by integrating a robust monocular depth estimation model to adapt RAFT-Stereo into a new framework.

Note that there are some contemporary works~\cite{bartolomei2024stereo,cheng2025monster,wen2025foundationstereo} that incorporate the depth foundation model into the recurrent stereo matching framework. They both use 3D convolutions upon a 4D cost volume to obtain initial disparity following~IGEV-Stereo~\cite{xu2023iterative} and ~\cite{bartolomei2024stereo,cheng2025monster} further get another refined initial disparity via alignment with the DEFOM depth. Our model differs in directly replacing the $\mathbf{0}$ initial disparity of the original RAFT-stereo with the  DEFOM depth and reforming the recurrent update process accordingly.

\subsection{Foundation Model for Depth Estimation} 
MiDaS~\cite{ranftl2020towards} pioneers zero-shot relative depth estimation (ZS-RDE), using a scale-shift-invariant loss to handle depth distribution inconsistencies in large-scale mixed datasets, achieving robust depth estimation across diverse scenes with CNNs. Later, DPT~\cite{ranftl2021vision} introduces a dense prediction head for the vision transformers, yielding finer-grained and more globally coherent depth predictions on large-scale training datasets.
MiDaS v3.1~\cite{birkl2023midas} conducts an in-depth and comprehensive evaluation of vision transformer backbones with varying architectures and pre-training methods, demonstrating that enhanced backbones consistently yield performance improvements. More recently, studies~\cite{depthanything, depth_anything_v2, ke2024repurposing} leverage powerful vision foundation models for ZS-RDE task, resulting in a significant enhancement of zero-shot generalization capabilities and ushering the field into a new era.

Among the public foundational models, Stable Diffusion (SD)~\cite{rombach2022sd}, trained on vast datasets for high-quality image generation, and DINOv2~\cite{oquab2023dinov2}, a general-purpose vision foundation model, are proven to have strong zero-shot generalization capabilities in both monocular and multi-view geometric estimation~\cite{el2024probing}.
Marigold~\cite{ke2024repurposing} fine-tunes pre-trained SD on limited synthetic depth data, achieving strong cross-dataset accuracy and capturing fine details. However, the diffusion process results in high inference costs and slow processing. Depth Anything~\cite{depthanything} employs DINOv2 as the encoder in the DPT architecture and proposes training a large powerful teacher model on labeled data, followed by distilling a student model on vast unlabeled real-world images to enhance its generalization performance. While being much faster, Depth Anything recovers fewer details than Marigold.

Building upon Depth Anything, Depth Anything V2~\cite{depth_anything_v2} reevaluates training data construction and notes that depth label noise from real captured datasets hinders the network's ability to generate detailed predictions.
It turns to use large-scale precise synthetic datasets to train a giant teacher model which subsequently distills student models on extensive unlabeled real data, yielding accurate and efficient depth foundation models.
Since Depth Anything V2 can rapidly infer detailed and accurate relative depth across various scenes, even including challenging areas such as reflective and transparent regions, we propose integrating its monocular depth cues for robust disparity estimation from stereo images.

\section{Methodology}
\label{sec:method}

\begin{figure*}[t]
    \centering{
    \includegraphics[width=0.95\linewidth]{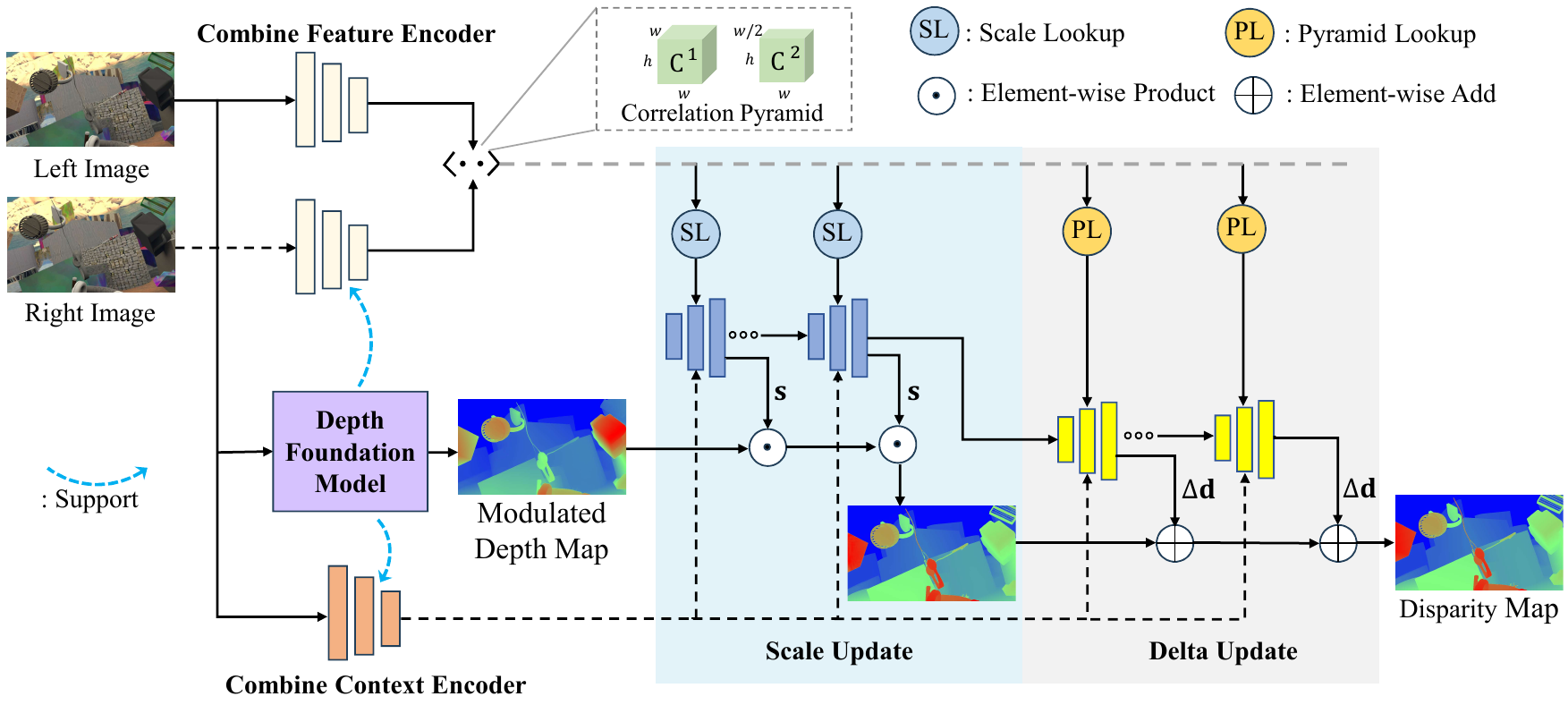}}
    \vspace{-10pt}
    \caption{Overview of our proposed DEFOM-Stereo. At first, we use a depth foundation model to augment the plain CNN encoders, thus obtaining more powerful feature and context encoders. Next, we insert the scale update module before the conventional delta update module. The SU starts from the modulated depth estimate from DEFOM and recurrently performs dense scale recovery with scaling lookup from correlation volume, while the DU continues to refine the result of SU with the pyramid lookup to obtain the final disparity map. }
    \label{fig:defom_stereo}
    \vspace{-10pt}
\end{figure*}

\subsection{Overall Framework}

Fig.~\ref{fig:defom_stereo} shows the overall framework of DEFOM-Stereo, which incorporates a depth foundation model, Depth Anything V2~\cite{depth_anything_v2} into the recurrent stereo model, RAFT-Stereo~\cite{lipson2021raft}. The framework consists of two main stages. The first stage mainly includes feature extraction, correlation pyramid construction, and recurrent disparity initialization. In this stage, we use the pre-trained DEFOM to enhance feature extraction and its predicted depth map to initialize disparity. In the next disparity update stage, we propose a scale update module to recover the accurate repeatedly dense scales for the disparities and then apply the conventional delta update to obtain finer details.

\subsection{Combined Feature Extraction}
\label{sec:ife}

There are two encoders in feature extraction. The first encoder is the feature encoder, which extracts the matching feature maps at $1/4$ resolution from the left and right images. Given the image size $H\times W$, we define $h=H/4$ and $w=W/4$ as the feature spatial size. The matching feature maps are used to construct the all-pair correlation volume, which provides matching cues for the recurrent update. 

Another is the context encoder, which is only applied on the left images to obtain multi-level context features at $1/4$, $1/8$, and $1/16$ resolutions. The context features are employed to initialize the update operator's hidden state and integrated into the GRU at each iteration of the update process, providing the monocular cue for the update process. In RAFT-Stereo~\cite{lipson2021raft}, the feature and context encoders are implemented with plain CNNs. 

We propose a simple but effective way to combine the pre-trained DEFOM and plain CNNs to construct improved feature extractors for stereo matching. We use the pre-trained backbone from the DEFOM, and initialize a new shared dense prediction transformer (DPT) to predict the ViT features for both the matching feature and context, as it is trainable compared with the existing DPT in the DEFOM, which has to be fixed for predicting depth.  

\textbf{Combined Feature Encoder.} We take the final fused feature map of the DPT and further, use a convolutional block to align it with the CNN feature map in the channel. After that, we simply add them as a combined feature map.

\textbf{Combined Context Encoder.} We take the feature maps from the $Reassemble_4$, $Reassemble_8$, and $Reassemble_{16}$ of the DPT as the ViT context maps. Similarly, we use another three convolutional blocks for channel alignment. Finally, we add the aligned ViT map and the CNN map to get the combined context map at different resolutions.

\begin{figure*}[t]
\centering
\input{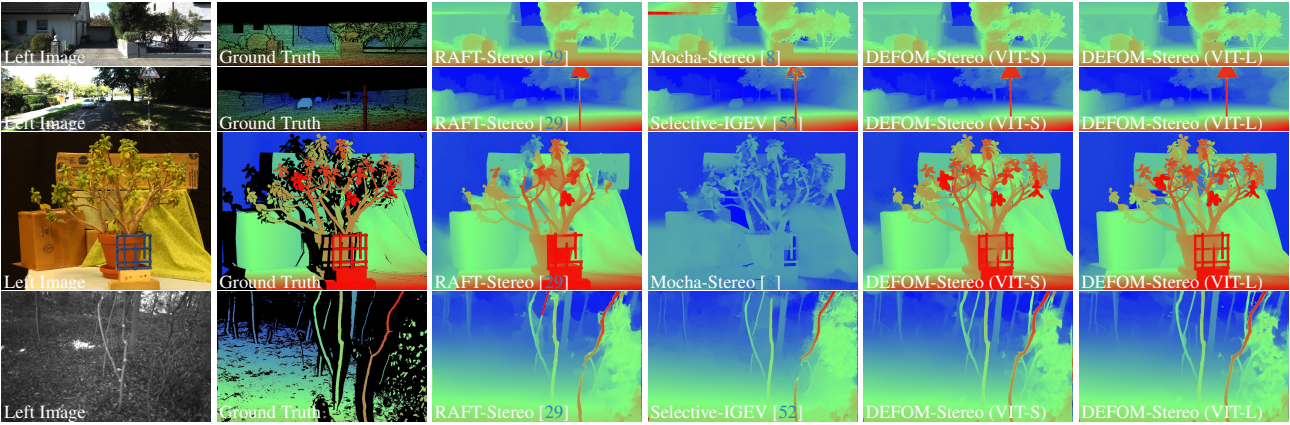}
\vspace{-8pt}
\caption{Zero-Shot Qualitative Comparison with RAFT-Stereo~\cite{lipson2021raft}, Selective-IGEV~\cite{wang2024selective} and Mocha-Stereo~\cite{chen2024mocha} on the four stereo datasets.  \textbf{Row 1:} KITTI 2012. \textbf{Row 2:} KITTI 2015. \textbf{Row 3:} Middlebury-full. \textbf{Row 4:} ETH3D. Best viewed in color and by zooming in.}
\label{fig:zeroshot}
\vspace{-12pt}
\end{figure*}

\subsection{Correlation Construction and Lookup}
\label{sec:ccl}
The section shows how to construct the all-pair correlation volume pyramid and perform correlation pyramid lookup. 
Given the extracted feature maps, $\mathbf{f}^l, \mathbf{f}^r$ ($\mathbf{f}\in \mathbb{R}^{c \times h \times w}$), the first level all-pairs correlation volume is computed as,
\begin{equation}
    \mathbf{C}^1_{ijk} = \sum_{h} \mathbf{f}^l_{hij} \cdot \mathbf{f}^r_{hik}, \mathbf{C}^1 \in \mathbb{R}^{h \times w \times w}.
    \label{eq:corr}
\end{equation}
Then, the multi-level correlation pyramid $\{\mathbf{C}^{l}\}$ can be obtained by repeatedly applying 1D average pooling at the last dimension to the upper correlation volume, and the last dimension will reduced by half each time. 

With the correlation pyramid created, the pyramid lookup (PL) is performed as follows. 
Given the current disparity estimate, the correlation values of it and its neighbors within a radius $r$ will be sampled from the correlation pyramid and concatenated into the retrieved correlation feature. 

RAFT-Stereo~\cite{lipson2021raft} uses a 4-level pyramid and sets the radius $r$ as $4$, so the maximum search range is $2^2\times 2^3\times 4=128$, which might be insufficient for the large disparity. In contrast, the proposed scale lookup does not have this limitation. As our scale update can provide approximated results, the delta update should focus on local details and we set the pyramid level as 2 following IGEV-Stereo~\cite{xu2023iterative}. 

\subsection{Delta Update}
\label{sec:du}

In the iterative motion estimation framework~\cite{teed2020raft, lipson2021raft}, the motion field is initialized with $\mathbf{0}$ and updated with a residual. For disparity, it  is updated in an additive manner, 
\begin{equation}
\mathbf{d}_{n}=\mathbf{d}_{n-1}+\Delta{\mathbf{d}}, 
\label{eq:du}
\end{equation}
where $\Delta{d}$ is a delta disparity map estimated by a multi-level convolutional Gated Recurrent Unit (ConvGRU). 

The ConvGRU at resolutions of $1/8$ or $1/16$ takes the context map at the same resolution and the reshaped hidden state at adjacent resolutions as input and updates its hidden state. These two ConvGRUs help to increase the receptive field of the update operation. The finest ConvGRU at $1/4$ resolution takes not only the context map at $1/4$ resolution and the reshaped hidden state at $1/8$ resolution, but also the encoded feature of the current disparity estimate and retrieved correlation feature map to update its hidden state. The hidden state $h_{n}$ at $1/4$ resolution is updated as follows, 
\begin{equation}
\begin{aligned}
x_n = & \; [\text{Encoder}_{c}(\mathbf{c}), \text{Encoder}_{d}(\mathbf{d}_{n-1}), \text{Up}^2(h_{n}^{1/8})]\\
z_n = & \;\sigma(\text{Conv}([h_{n-1}, x_n], W_z) + c_z), \\
r_n = & \;\sigma(\text{Conv}([h_{n-1}, x_n], W_r) + c_r), \\
\Tilde{h}_n = & \,\tanh(\text{Conv}([r_n \odot h_{n-1}, x_n], W_h) + c_h), \\
h_{n} = & \;(1-z_n) \odot h_{n-1} + z_n \odot \Tilde{h}_n,
\end{aligned}
\label{eq:gru}
\end{equation}
where $c_z$, $c_r$, $c_h$ are context features generated from the context network. $h_{n}^{1/8}$ is the hidden state at $1/8$ resolution. $\mathbf{c}$ is the retrieved correlation feature map from the correlation volume pyramid with the previous disparity estimate $\mathbf{d}_{n-1}$. 

Finally, the updated $h_{n}$ will be used to predict the delta disparity map $\Delta{\mathbf{d}}$ in Eqn.~\ref{eq:du} with two convolutional layers. As the disparity estimate is at $1/4$ resolution, $h_{n}$ also predicts a convex weight map, so that $\mathbf{d}_{n}$ can be upsampled to the original resolution with the convex combination~\cite{teed2020raft}.

\subsection{Monocular Depth Initialization}
\label{sec:di}

We first propose to utilize the predicted depth map from the depth foundation model to initialize the recurrent disparity map. The depth foundation model usually outputs the relative depth, which has unknown scales and shifts to the disparity. 
To make reasonable initialization, we should convert the predicted depth proportional to the true disparity. Consider that the disparity amplitude of an image is usually proportional to its width. Suppose that the depth map predicted by the depth model is $\mathbf{z}$, we obtain the initialized disparity map with the following formula, 
\begin{equation}
\mathbf{d}_0= \frac{\eta w*\mathbf{z}}{\max(\mathbf{z})} + \epsilon. 
\label{eq:di}
\end{equation}
where $\eta$ is the controlling ratio and $\epsilon$ is a tiny positive bias to make scale update work when the predicted depth is zero.

\subsection{Scale Update}
\label{sec:su}
As the initialized disparity map is scale-uncertain and inconsistent, we first use the scale update module to obtain an overall scale-certain and consistent disparity map recurrently. Supposed the disparity map in the last iteration is $\mathbf{d}_{n-1}$, the ConvGRU for SU would predict a dense scale map $\mathbf{s}$, the next disparity map is updated by, 
\begin{equation}
\mathbf{d}_{n}=\mathbf{s} \cdot \mathbf{d}_{n-1}.
\label{eq:su}
\end{equation}

It is insufficient to feed the ConvGRU with the sampled correlation feature from the pyramid lookup to recover the scale, as its search range is limited. To this end, we design the scale lookup (SL) that retrieves correlation in a scaling manner. The SL is only applied to the finest correlation volume. Given the disparity estimation $\mathbf{d}_n$, we multiply it with a series of pre-defined scale factors {$s^m$}, then the correlation value corresponding to the scaled disparity $s^m\mathbf{d}_n$, $\mathbf{C}^1(s^m\mathbf{d}_n)$ will be sampled. To, make the sampled correlation map more robust to local noise, the correlation of two nearby pixels, $\mathbf{C}^1(s^m\mathbf{d}_n-1)$ and $\mathbf{C}^1(s^m\mathbf{d}_n+1)$ will also be retrieved. 
To enable the SL to reach the whole image pixels, the ratio $\eta$ in Eqn.~\ref{eq:di} is set to $1/2$, and the scale factors are set to $\{1, 2, 4, 6, 8, 10, 12, 16\}/8$. In total, 24 correlation values are retrieved in the scale lookup.

\subsection{Loss Function}
Suppose that the total number of update iterations is $N$ and the set of predicted disparity maps are $\{\mathbf{d}_1, ..., \mathbf{d}_N \}$. We follow~\cite{teed2020raft, lipson2021raft} to supervise those predictions in exponentially increasing weights with the ground truth disparity $\mathbf{d}_{gt}$. Tho loss is computed as, 
\begin{equation}
    \mathcal{L} = \sum_{i=n}^{N} \gamma^{N-n} ||\mathbf{d}_{gt} - \mathbf{d}_n||_1, \qquad \text{where } \gamma=0.9.
\label{eq:loss}
\end{equation}

\section{Experiments}
\label{sec:exp}
In this section, we verify the effectiveness of our model via synthetic-to-real zero-shot generalization, ablation study and evaluation on popular online stereo benchmarks. The main ablation study shows the effectiveness and complexities of the proposed components and the additional ablation study of the combined encoders' design choices and the iterations of scale updates will be presented on supplementary. 


\begin{table}
\centering
\resizebox{1.00\columnwidth}{!}{
  \begin{tabular}{l|cc|ccc|c}
    \toprule
    \multirow{2}{*}{Methods} &\multicolumn{2}{c|}{KITTI} &\multicolumn{3}{c|}{Middlebury} &\multirow{2}{*}{ETH3D}\\
    & 2012 & 2015 & full & half & quarter & \\
    \midrule
    DSMNet~
    \cite{zhang2020domain} & 6.2 & 6.5 & 21.8 & 13.8 & 8.1 & 6.2 \\
    RAFT-Stereo~\cite{lipson2021raft} & 4.35 & {5.74} & {18.33} & {12.59} & {9.36} & {3.28} \\ 
    DLNR~\cite{zhao2023high} & 9.09 & 16.05 & 16.78 & 9.82 & 7.82 & 22.99 \\
    IGEV-Stereo~\cite{xu2023iterative}  & 5.14 & 6.03 & 36.41 & 13.36 & 8.82 & 4.05 \\
    Selective-RAFT~\cite{wang2024selective} & 5.19 & 6.68 & 33.98 & 13.25 & 8.66 & 4.36 \\
    Selective-IGEV~\cite{wang2024selective} & 5.65 & 6.05 & 35.82 & 13.27 & 9.82 & 6.07 \\
    NMRF-Stereo~\cite{guan2024neural} & \underline{4.23} & 5.52 & 44.30 & 15.88 & 7.49 & 3.80 \\
    Mocha-Stereo~\cite{chen2024mocha}  & 4.85 & 5.93 & 35.34 & 11.49 & 7.39 & 3.89 \\
    \hline
    DEFOM-Stereo (ViT-S) & 4.29 & \underline{5.29} & \underline{14.70} & \underline{6.76} & \underline{6.38} & \underline{2.61} \\
    DEFOM-Stereo (ViT-L) & \textbf{3.76} & \textbf{4.99} & \textbf{11.95} & \textbf{5.91} & \textbf{5.65} & \textbf{2.35} \\
    \bottomrule
  \end{tabular}
}
  \vspace{-2mm}
  \caption{\textbf{Synthetic to real generalization}. All models are trained on Scene Flow. Standard thresholding error rates are used:  3-pixel for KITTI, 2-pixel for Middlebury, and 1-pixel for ETH3D.}
  \label{tab:zero-shot}
  \vspace{-10pt}
\end{table}

\subsection{Implementation Details}

We implement our DEFOM-Stereo using the PyTorch framework on NVIDIA RTX 4090 GPUs. We use a simplified REAF-Stereo that reduces the correlation pyramid level from 4 to 2 as our baseline model and add the proposed components to construct DEFOM-Stereo. 
For all training, we use the AdamW optimizer, a one-cycle learning rate schedule with a learning rate of 2e-4 and a batch size of 8. The total number of update iterations (SU+DU) is set to 18 in training and 32 in evaluation. The number of SU is fixed to 8 during both training and evaluation. $\epsilon$ in Eqn.~\ref{eq:di} is set to 0.05.

\subsection{Zero-shot Generalization}
\label{sec:pc}

We first pre-train our model on Scene Flow for 200k with a crop size of $320 \times 736$.  We follow~\cite{zhang2020domain, lipson2021raft} to examine the zero-shot generalization ability of our model on the trainsets of four realistic datasets, and for Middlebury, we consider the three available resolutions. 
For Middlebury and ETH3D, we count all valid pixels instead of only those in non-occluded regions, as this allows for a more comprehensive test. For other methods, we perform such evaluation using the provided pre-trained model on Scene Flow. The results are shown in Tab.~\ref{tab:zero-shot}. We found that despite later recurrent optimation-based methods~\cite{zhao2023high, xu2023iterative, wang2024selective, chen2024mocha} have improved on pre-training fitting on Scene Flow compared with RAFT-Stereo, fail to make comprehensive progress on zero-shot evaluation. For example, DLNR~\cite{zhao2023high}, shows a clear performance gain on Middlebury with different resolutions, but significantly increases the error rate on KITTI and ETH3D. For the recent neural Markov random field-based model, NMRF-Stereo~\cite{guan2024neural}, progress can be seen only in driving datasets KITTI 2012 and 2015, and Middlebury-quarter.

For DEFOM-Stereo, we present two variants with two different ViT backbone sizes, large (ViT-L) and small (ViT-S). Both model variants achieve improvement in all these realistic datasets. In particular, compared with RAFT-Stereo, DEFOM-Stereo (ViT-L) reduces the error rates, by about $13\%$ on KITTI, by $28\%$ on ETH3D, and by over a half on Middlebury. Furthermore, on the high-resolution Middlebury-full, DEFOM-Stereo can even reduce the error rate by over $29\%$  compared with the previous best model (DLNR), demonstrating the strong generalization ability with high-resolution even pre-trained at low resolution.

\begin{table*}[t]
  \centering
\resizebox{2.0\columnwidth}{!}{
  \begin{tabular}{l|cccc|cc|c|c|c|c|cc}
    \toprule
    \multirow{2}{*}{Models} & \multicolumn{4}{c|}{Proposed Modules} & \multicolumn{2}{c|}{Scene Flow} & {KITTI 2012} & {KITTI 2015} & {Middlebury-half} & {ETH3D}   & \multirow{2}{*}{Params.  (M)} &  \multirow{2}{*}{Time (s)} \\
     & CCE & CFE & DI & SU & EPE & Bad 1.0 & Bad 3.0 & Bad 3.0 & Bad 2.0 & Bad 1.0 &   & \\
    \midrule
      Baseline          &            &            &            &            & 0.56 & 6.66 & 4.65 & 5.57 & 10.67 & 3.45 & 11.11 & 0.222\textdagger \\
      \hline
      +CCE              & \checkmark &            &            &            & 0.49 & 6.08 & 4.40 & 5.84 & 8.42 & 2.82 & 12.10 & 0.242 \\
      +CFE              &            & \checkmark &            &            & 0.50 & 6.17 & 4.13 & 5.53 & 10.45 & 2.83 & 13.89 & 0.243 \\
      +CCE+CFE          & \checkmark & \checkmark &            &            & 0.49 & 5.95 & 4.02 & 5.75 & 8.31 & 2.53 & 14.11 & 0.246 \\
      +DI               &            &            & \checkmark &            & 0.57 & 6.74 & 4.57 & 5.63 & 12.40 & 2.77 & 11.11 & 0.242 \\
      +DI+SU            &            &            & \checkmark & \checkmark & 0.50 & 6.03 & 4.15 & 5.12 & 8.15 & 2.67 & 15.51 & 0.244 \\
      Full Model (ViT-S) & \checkmark & \checkmark & \checkmark & \checkmark & 0.46 & 5.57 & 4.29 & 5.29 & 6.76 & 2.61 & 18.51 & 0.255 \\
      \hline
      Full Model (ViT-L) & \checkmark & \checkmark & \checkmark & \checkmark & 0.42 & 5.10 & 3.76 & 4.99 & 5.91 & 2.35 & 47.30 & 0.316 \\
   \bottomrule
  \end{tabular}
  }
  \caption{\textbf{Ablation study of proposed networks on the Scene Flow test set and zero-shot generation.} The baseline is RAFT-Stereo with two levels of correlation pyramids. The parameters counted here are the trainable ones. The time is the inference time for 960$\times$540 inputs. \textdagger \textit{We found that pre-defining the neighbor sampling indexes within the search radius can significantly accelerate the inference instead of repeatedly defining them in every lookup as RAFT-Stereo's implementation.  We also apply this trick to the baseline, otherwise, its inference time would be 0.329s. }}
  \label{tab:ablation}
\end{table*}

\subsection{Ablation Study}
\label{sec:as}

In this section, we verify the effectiveness of the proposed components via a main ablation study. We mainly use the ViT-S as the ViT backbone for our model and train all the variants on Scene Flow for 200k steps. For accuracy comparison, both the in-domain test and zero-shot generalization evaluation are presented. We also list trainable parameters and the inference time of the model variants for comparing computational complexities. The results are shown in Tab.~\ref{tab:ablation}.

\textbf{Effecitveness of combined encoders.} 
Compared with the baseline, the combined context encoder (CCE) and the combined feature encoder (CFE) achieves about $10\%$ improvement on Scene Flow, while CCE performs slightly better than CFE. Their combination does not give much additional gain. Clear improvement appears on KITTI 2012, Middlebury, and ETH3D, but KITTI 2015. 

\textbf{Effecitveness of depth initialization.} 
Initializing the disparity map with the modulated depth from DEFOM instead of zeros does not improve the model's performance on in-domain fitting. Nevertheless, DI achieves better generalization results on KITTI 2012, Middlebury, and ETH3D. 

\textbf{Effecitveness of scale update.} 
When incorporating the scale update with depth initialization (+DI+SU), there is around $10\%$ improvement on Scene Flow. Besides, apparent progress is also obtained in the generationalization evaluation of the four realistic datasets. Noteworthy, the Bad 2.0 of Middlebury is reduced by over $50\%$. 

\textbf{Effecitveness of all components.} 
By integrating all the proposed modules into our complete model, we observe further improvements on the Scene Flow and Middlebury datasets. However, some slight performance drops are observed when compared to individual components on other datasets. For instance, on KITTI 2015, the full model (ViT-S) slightly underperforms the combination of depth initialization and scale update. Nevertheless, the full model demonstrates better overall performance. Additionally, using a larger ViT backbone, ViT-L, further enhances performance.

\textbf{Trainbale Parameters.}  As we fixed the DEFOM, the new trainable parameters mainly come from the new DPT for CCE and CFE, and the ConvGRU for SU. using CCE and CFE meanwhile increases 2M parameters (+$18\%$), using SU increases 5.4M parameters (+$49\%$),  and the full model (ViT-S) increases 7.4M parameters (+$67\%$). The full model (ViT-L) has quadrupled the parameters, as the channels of the new DPT for CCE and CFE are defined to be proportional to those of the ViT backbone, following the fixed DPT of DEFOM for predicting depth. 

\textbf{Inference times.}  The increase in inference time is not as significant as the growth in model parameters, as the majority of the inference time is spent on recurrent update iterations. The total number of iterations in our model is set to match that of RAFT-Stereo. The proposed components contribute to a modest $10\%$ increase in inference time individually, primarily due to the operation of DEFOM. The full model (ViT-S) results in a $15\%$ increase in inference time, while the larger full model (ViT-L) sees a $42\%$ increase.

\def \rank#1{$^{\textcolor{red}{#1}}$ }
\begin{table*}[t]
\centering
\resizebox{2.08\columnwidth}{!}{
\begin{tabular}{l|cccccc|cccccc|c}
\toprule
\multirow{3}{*}{Method} & \multicolumn{6}{c|}{KITTI-2012}                                                             & \multicolumn{6}{c|}{KITTI-2015}               & \multirow{3}{*}{\begin{tabular}[c]{@{}c@{}}\\Run-time\\ (s)\end{tabular}} \\
& \multicolumn{3}{c}{non-occluded}    & \multicolumn{3}{c|}{all}       & \multicolumn{3}{c}{non-occluded}    & \multicolumn{3}{c|}{all}   \\
\cmidrule(lr){2-4} \cmidrule(lr){5-7} \cmidrule(lr){8-10} \cmidrule(lr){11-13}
                                            & Bad 2.0 & Bad 3.0 & AvgErr & Bad 2.0 & Bad 3.0 & AvgErr & D1-bg & D1-fg & D1-all & D1-bg & D1-fg & D1-all &  \\ 
\midrule
ACVNet~\cite{xu2022attention}               & 1.83  & 1.13  & 0.4  & 2.34   & 1.47  & 0.5  
& 1.26  & 2.84  & 1.52  & 1.37  & 3.07  & 1.65  & 0.2  \\
PCWNet~\cite{shen2022pcw}                   & 1.69  & 1.04  & 0.4  & 2.18   & 1.37  & 0.5  & 1.26  & 2.93  & 1.53  & 1.37  & 3.16  & 1.67  & 0.44  \\
LaC + GANet~\cite{liu2022local}             & 1.72  & 1.05  & 0.4  & 2.26   & 1.42  & 0.5  & 1.26  & 2.64  & 1.49  & 1.44  & 2.83  & 1.67  & 1.8  \\
CREStereo~\cite{li2022practical}            & 1.72  & 1.14  & 0.4  & 2.18   & 1.46  & 0.5  & 1.33  & 2.60  & 1.54  & 1.45  & 2.86  & 1.69  & 0.40  \\
IGEV-Stereo~\cite{xu2023iterative}          & 1.71  & 1.12  & 0.4  & 2.17   & 1.44  & 0.4  & 1.27  & 2.62  & 1.49  & 1.38  & 2.67  & 1.59  & 0.18  \\
MC-Stereo~\cite{feng2024mc}                 & 1.55  & 1.04  & 0.4  & 1.99   & 1.34  & 0.4  & 1.24  & 2.55  & 1.46  & 1.36  & 2.51  & 1.55  & 0.40  \\
DN+ACVNet ~\cite{zhang2024exploring}        & 1.64  & 1.02  & 0.4  & 2.20   & 1.41  & 0.5  & 1.21  & 2.62  & 1.44  & 1.32  & 2.95  & 1.60  & 0.24  \\
Selective-IGEV~\cite{wang2024selective}     & 1.59  & 1.07  & 0.4  & 2.05   & 1.38  & 0.4  & 1.22  & 2.55  & 1.44  & 1.33  & 2.61  & 1.55  & 0.24  \\
NMRF-Stereo~\cite{guan2024neural}           & 1.59  & 1.01 & 0.4   & 2.07   & 1.35  & 0.4  & 1.18  & 2.90  & 1.46  & 1.28  & 3.07  & 1.57  & 0.09  \\
LoS~\cite{li2024local}                      & 1.69  & 1.10  & 0.4  & 2.12   & 1.38  & 0.4  & 1.29  & 2.66  & 1.52  & 1.42  & 2.81  & 1.65  & 0.19 \\
GANet+ADL~\cite{xu2024adaptive}             & 1.52  & 0.98  & 0.4   & 2.01  & 1.29  & 0.5  & 1.24  & \bf2.18 & 1.40  & 1.38  & 2.38  & 1.55  & 0.67 \\
MoCha-Stereo~\cite{chen2024mocha}           & 1.64  & 1.06  & 0.4  & 2.07   & 1.36  & 0.4  & 1.24  & 2.42  & 1.44  & 1.36  & 2.43  & 1.53  & 0.33  \\
IGEV++~\cite{xu2024igev++}                  & 1.56  & 1.04  & 0.4  & 2.03   & 1.36  & 0.4  & 1.20  & 2.54  & 1.42  & 1.31  & 2.54  & 1.51  & 0.28  \\
StereoBase~\cite{guo2023openstereo}         & 1.54  & 1.00  & 0.4  & 1.95   & 1.26  & 0.4  & 1.17  &  \underline{2.23}  & \underline{1.35}  & 1.28  & \underline{2.26}  & \underline{1.44}  & 0.29  \\
ViTAStereo~\cite{liu2024playing}            & \underline{1.46}  & \bf 0.93  & 0.4  & \underline{1.80}   & \bf 1.16  & 0.4  & \bf 1.12  & 2.90  & 1.41  & \bf 1.21  & 2.99  & 1.50  & 0.22  \\
\midrule
DEFOM-Stereo (Ours)                         & \bf 1.43 & \underline{0.94} & \textbf{0.3}\rank{1} & \bf 1.79  & \underline{1.18} & 0.4
& \underline{1.15} & {2.24} & \textbf{1.33}\rank{1} &\underline{1.25} & \textbf{2.23}\rank{1} & \textbf{1.41}\rank{1} & 0.30  \\ 
\bottomrule                  
\end{tabular} 
}
\vspace{-5pt}
\caption{\textbf{Quantitative evaluation on KITTI 2012 and KITTI 2015.} \rank{1} $1^{st}$-ranking metric on the leaderboards at the time of writing. }
\label{tab:kitti}
\end{table*}

\begin{table*}[t]
\centering
\resizebox{1.90\columnwidth}{!}{
\begin{tabular}{l|cccccc|cccccc}
\toprule
\multirow{2}{*}{Method} & \multicolumn{6}{c|}{Middlebury}            & \multicolumn{6}{c}{ETH3D}   \\
& \multicolumn{3}{c}{non-occluded}    & \multicolumn{3}{c|}{all}       & \multicolumn{3}{c}{non-occluded}    & \multicolumn{3}{c}{all}   \\
\cmidrule(lr){2-4} \cmidrule(lr){5-7} \cmidrule(lr){8-10} \cmidrule(lr){11-13} 
                                    & Bad 2.0 & AvgErr & RMS & Bad 2.0 & AvgErr & RMS 
                                    & Bad 1.0 & AvgErr & RMS & Bad 1.0 & AvgErr & RMS\\
\midrule
HITNet~\cite{tankovich2021hitnet}        & 6.46 & 1.71 & 9.97 & 12.8 & 3.29 & 14.5 & 2.79 & 0.20 & 0.46 & 3.11 & 0.22 & 0.55 \\
RAFT-Stereo~\cite{lipson2021raft}          & 4.74 & 1.27 & 8.41 & 9.37 & 2.71 & 12.6 & 2.44 & 0.18 & 0.36 & 2.60 & 0.19 & 0.42 \\
CREStereo~\cite{li2022practical}         & 3.71 & 1.15 & 7.70 & 8.13 & 2.10 & 10.5 & 0.98 & 0.13 & 0.28 & 1.09 & 0.14 & 0.31 \\
CroCo-Stereo~\cite{weinzaepfel2023croco} & 7.29 & 1.76 & 8.91 & 11.1 & 2.36 & 10.6 & 0.99 & 0.14 & 0.30 & 1.14 & 0.15 & 0.35 \\
GMStereo~\cite{xu2023unifying}           & 7.14 & 1.31 & \underline{6.45} & 11.7 & 1.89 & \underline{8.03} & 1.83 & 0.19 & 0.38 & 2.07 & 0.21 & 0.44 \\
IGEV-Stereo~\cite{xu2023iterative}       & 4.83 & 2.89 & 12.8 & 8.16 & 3.64 & 15.1 & 1.12 & 0.14 & 0.34 & 1.51 & 0.20 & 0.86 \\
DLNR~\cite{zhao2023high}                 & 3.20 & 1.06 & 7.78 & 6.98 & 1.91 & 10.2 &  -   &  -   &  -   &  -   &  -   &  -   \\
LoS~\cite{li2024local}                   & 4.20 & 1.12 & 6.99 & 8.03 & 1.75 & 8.78 & \underline{0.91} & 0.14 & 0.31 & \underline{1.03} & \underline{0.15} & \underline{0.34} \\
IGEV++~\cite{xu2024igev++}               & 3.23 & 0.97 & 7.23 & 7.19 & 1.83 & 10.2 & 1.14 & 0.13 & 0.34 & 1.58 & 0.19 & 0.74 \\
Selective-IGEV~\cite{wang2024selective}  & \underline{2.51} & \underline{0.91} & 7.26 & \underline{6.04} & \underline{1.54} & 9.26 & 1.23 & \underline{0.12} & \underline{0.29} & 1.56 & \underline{0.15} & 0.57 \\
DEFOM-Stereo (Ours)                      & \textbf{2.39}\rank{2} & \textbf{0.79}\rank{1} & \textbf{5.81}\rank{1} & \textbf{5.02}\rank{1} &\textbf{1.27}\rank{1} & \textbf{7.73}\rank{1} 
                                         & \textbf{0.70}\rank{4}  & \textbf{0.11}\rank{3} & \textbf{0.22}\rank{2} & \textbf{0.78}\rank{2} &\textbf{0.11}\rank{1} & \textbf{0.26}\rank{1}  \\
\bottomrule
\end{tabular}
}
\vspace{-5pt}
\caption{\textbf{Quantitative evaluation on ETH3D and Middlebury benchmarks.} The superscript number on the metric of our model represents its ranking on the leaderboards at the time of writing.}
\label{tab:middlebury_eth3d}
\end{table*}

\subsection{Benchmark Comparisons}
\label{sec:bc}

In this section, we present the results of our models against the previous SOTA methods on the typical stereo leaderboards of KITTI 2012~\cite{geiger2012we}, KITTI 2015~\cite{menze2015object}, Middlebury~\cite{scharstein2014high}, and ETH3D~\cite{schops2017multi}. 

\textbf{KITTI.}  For submitting the benchmarks of KITTI, we first finetune the Scene Flow pre-trained model 50k on the mixed dataset of KITTI 2012~\cite{geiger2012we}, KITTI 2015~\cite{menze2015object} and virtual KITTI 2~\cite{vkitti}, where the samples of KITTI 2012 and KITTI 2015 are augmented to take account for $50\%$ of the mixed dataset. On the official leaderboards, our model obtains many $1^{st}$ ranking metrics among all the submissions at the time of paper writing, including Avg-Noc, Avg-All(Reflective), and 2-Out-All(Reflective) for KITTI 2012 and D1-fg, D1-all and D1-all(Noc) for KITTI. 
The results compared with public SOTA methods are shown in Tab.~\ref{tab:kitti}. Compared with StereoBase~\cite{guo2023openstereo} and ViTAStereo~\cite{liu2024playing}, which perform best among the public methods, our model still shows clear advantages. 
For ViTAStereo, while its performance on KITTI 2012 is comparable, it underperforms our model overall on KITTI 2015, particularly in the foreground region. For StereoBase, our model is slightly better than it on KITTI 2015 but outperforms it on KITTI 2012 by a clear margin, for example, $8.2\%$ reduction of Bad 2.0 (all).

\textbf{Middlebury.} Following previous work~\cite{li2022practical, xu2023unifying, wang2024selective}, we first finetune the Scene Flow pre-trained model on a mixed dataset fo Tartan Air~\cite{wang2020tartanair}, CREStereo Dataset~\cite{li2022practical}, Scene Flow~\cite{sceneflow}, Falling Things~\cite{tremblay2018falling}, InStereo2k~\cite{bao2020instereo2k}, CARLA HR-VS~\cite{yang2019hierarchical} and Middlebury~\cite{scharstein2014high} datasets with a crop size of $384 \times 512$ for 200k steps. Next, we perform finetuning on a combination  of CREStereo Dataset~\cite{li2022practical}, Falling Things~\cite{tremblay2018falling}, InStereo2k~\cite{bao2020instereo2k}, CARLA HR-VS~\cite{yang2019hierarchical} and Middlebury~\cite{scharstein2014high} datasets with a crop size of $512 \times 768$ for 100k steps. Our model delivers outstanding performance on the Middlebury benchmark: DEFOM-Stereo ranks first in 7 out of 10 accuracy metrics when evaluated on all valid pixels, and in 3 out of 10 metrics when evaluated on non-occluded valid pixels. The comparison with public methods is shown on the left of Tab.~\ref{tab:middlebury_eth3d}. Quantitatively, our model outperforms the best-published method, Selective-IGEV~\cite{wang2024selective} by $4.8\%$ on Bad 2.0 (noc) and $16.9\%$ on Bad 2.0 (all). The gap in performance gain between evaluation on non-occluded and all valid pixels suggests our model brings more improvement on occluded regions than non-occluded ones.

\textbf{ETH3D.} Similarly, in finetuning for ETH3D benckmark submisison, we follow CREStereo~\cite{li2022practical}, GMStereo~\cite{xu2023unifying} and Selective-Stereo~\cite{wang2024selective}. The first finetuning was performed on the mixture of Tartan Air~\cite{wang2020tartanair}, CREStereo Dataset~\cite{li2022practical}, Scene Flow~\cite{sceneflow}, Sintel Stereo~\cite{butler2012naturalistic}, InStereo2k~\cite{bao2020instereo2k} and ETH3D~\cite{schops2017multi} with a crop size of $384 \times 512$ for 300k steps. Then, another finetuning was only performed on CREStereo Dataset~\cite{li2022practical}, InStereo2k~\cite{bao2020instereo2k} and ETH3D~\cite{schops2017multi} datasets for 90k steps. Still, our model obtains the top performance on the ETH3D leaderboard and has several $1^{st}$-ranking metrics when evaluated with all valid masks. The results of previous methods and ours are listed on the right of Tab.~\ref{tab:middlebury_eth3d}. Compared with LoS~\cite{li2024local}, which has the best Bad 1.0 among the published methods, our model reduces Bad 1.0 (noc) by $23.1\%$, and Bad 1.0 (all) by $24.3\%$, a noteworthy progress.

\subsection{Robust Vision Challenge}

\begin{table*}[t]
\centering
\resizebox{2.1\columnwidth}{!}{
\begin{tabular}{l|cccccc|cccc|cccc}
\toprule
\multirow{2}{*}{Method} & \multicolumn{6}{c|}{KITTI-2015} & \multicolumn{4}{c|}{Middlebury}  & \multicolumn{4}{c}{ETH3D}   \\
& \multicolumn{3}{c}{non-occluded}    & \multicolumn{3}{c|}{all}  & \multicolumn{2}{c}{non-occluded}    & \multicolumn{2}{c|}{all}       & \multicolumn{2}{c}{non-occluded}    & \multicolumn{2}{c}{all}   \\
\cmidrule(lr){2-4} \cmidrule(lr){5-7} \cmidrule(lr){8-9} \cmidrule(lr){10-11} \cmidrule(lr){12-13} \cmidrule(lr){14-15} 
& D1-bg & D1-fg & D1-all & D1-bg & D1-fg & D1-all & Bad 2.0 & AvgErr & Bad 2.0 & AvgErr  & Bad 1.0 & AvgErr & Bad 1.0 & AvgErr \\
\midrule
\multicolumn{15}{c}{\href{http://robustvision.net/rvc2020.php}{Robust Vision Challenge 2018}} \\
\hline
DN-CSS\_ROB \cite{ilg2018occlusions}                     & 2.23 & 4.96 & 2.68 & 2.39 & 5.71 & 2.94   & 22.8 & 4.04 & 28.3 & 5.48   & 2.69 & 0.22 & 3.00 & 0.24  \\
iResNet\_ROB \cite{liang2018learning}                    & 2.10 & 3.96 & 2.40 & 2.27 & 4.89 & 2.71   & 24.8 & 4.51 & 31.7 & 6.56   & 4.23 & 0.25 & 4.67 & 0.27  \\

\midrule
\multicolumn{15}{c}{\href{http://robustvision.net/rvc2020.php}{Robust Vision Challenge 2020}} \\
\hline
NLCA\_NET\_v2\_RVC~\cite{rao2022rethinking}              & 1.36 & 3.49 & 1.71 & \underline{1.51} & 3.97 & 1.92   & 10.4 & 3.89 & 16.4 & 5.60   & 3.84 & 0.27 & 4.11 & 0.29 \\
CFNet\_RVC~\cite{shen2021cfnet}                          & 1.50 & 3.03 & 1.76 & 1.65 & 3.53 & 1.96   & 10.1 & 3.49 & 16.1 & 5.07   & 3.31 & 0.24 & 3.70 & 0.26 \\

\midrule
\multicolumn{15}{c}{\href{http://robustvision.net/}{Robust Vision Challenge 2022}} \\
\hline
iRaftStereo\_RVC~\cite{lipson2021raft,jiang2022improved} & 1.76 & 2.94 & 1.95 & 1.88 & \underline{3.03} & 2.07   & 8.07 & 1.71 & 13.3 & 2.90   & 1.62 & 0.16 & 1.88 & 0.17 \\
CREStereo++\_RVC~\cite{jing2023uncertainty}              & 1.43 & 3.36 & 1.75 & 1.55 & 3.53 & 1.88   & \underline{4.68} & 1.28 & 9.46 & \underline{2.20}   & 1.59 & 0.15 & 1.70 & 0.16  \\

\midrule
\multicolumn{15}{c}{More Recently} \\
\hline

UCFNet\_RVC~\cite{shen2023digging}                       & \underline{1.41} & \underline{2.93} & \underline{1.66} & {1.57} & 3.33 & 1.86   & 10.7 & 3.74 & 16.7 & 5.96   & 3.09 & 0.24 & 3.37 & 0.25 \\
LoS\_RVC~\cite{li2024local}                              & 1.46 & 2.95 & 1.71 & 1.58 & 3.08 & \underline{1.83}   & 5.14 & \underline{1.57} & \underline{9.30} & 2.36   & \underline{1.26} & \textbf{0.13} & \underline{1.47} & \underline{0.14} \\
DEFOM-Stereo\_RVC                                        & \textbf{1.32} & \textbf{2.67} & \textbf{1.54} & \textbf{1.42} & \textbf{2.68} & \textbf{1.63}   & \textbf{3.28} & \textbf{0.97} & \textbf{6.90} & \textbf{1.61}  & \textbf{0.98} & \textbf{0.13} & \textbf{1.09} & \textbf{0.13} \\
\bottomrule
\end{tabular}
}
\vspace{-5pt}
\caption{\textbf{Robust Vision Challenge.} Comparison with winners and runner-ups of the Robust Vision Challenges 2018, 2020, and 2022, and more recent RVC models, UCFNet\_RVC~\cite{shen2023digging} and LoS\_RVC~\cite{li2024local}.}
\label{tab:rvc}
\end{table*}

In this section, we examine our model under the setting of the \href{http://robustvision.net}{Robust Vision Challenge}. Different from the previous section which focuses on the performance of only a specific domain, this section will compare models' performance across many domains with the same model parameters. The three stereo benchmarks in RVC are KITTI 2015~\cite{menze2015object}, Middlebury~\cite{scharstein2014high}, and ETH3D~\cite{schops2017multi}. The RVC evaluation reveals the joint generalization capability of models. 

To provide a lightweight RVC model, we use the small ViT backbone, \ie, ViT-S, instead of the ViT-L for individual benchmarks. Following the training strategy for individual benchmarks in Sec.~\ref{sec:bc}, we use a mixture of synthetic and realistic datasets for finetuning the pre-trained model on Scene Flow~\cite{sceneflow}. Specifically, we perform the first finetuning for 200k on a mixture of synthetic datasets, including Tartan Air~\cite{wang2020tartanair}, CREStereo Dataset~\cite{li2022practical}, Scene Flow~\cite{sceneflow}, Falling Things~\cite{tremblay2018falling}, CARLA HR-VS~\cite{yang2019hierarchical}, Sintel Stereo~\cite{butler2012naturalistic},  virtual KITTI 2~\cite{vkitti}, IRS~\cite{wang2021irs} and 3D Ken Burns~\cite{niklaus20193d}. Next, we add the training split of realistic datasets, including KITTI 2012~\cite{geiger2012we}, KITTI 2015~\cite{menze2015object}, Middlebury~\cite{scharstein2014high}, ETH3D~\cite{schops2017multi}, InStereo2k~\cite{bao2020instereo2k} and Booster dataset~\cite{zamaramirez2022booster} for additional 100k finetuning. 
Following LoS\_RVC~\cite{li2024local}, we augment the KITTI 2012 and  KITTI 2015 to make up half of the mixed dataset in the final 20k fintuning.  During all the finetuning, the batch size and input crop size are set to 8 and $384 \times 768$. 

We compare our RVC model with previous RVC models in Tab~\ref{tab:rvc}. The previous models include the winners and runners-up of the Robust Vision Challenges in 2018, 2020, and 2022, and two more recent RVC models. Our DEFOM-Stereo\_RVC achieves the best results among all RVC models on three benchmarks and shows a significant performance gap over previous models on the KITTI 2015 and Middlebury. For example, DEFOM-Stereo\_RVC outperforms the next-best CREStereo++\_RVC and LoS\_RVC, by around 25\% on the listed four typical metrics on Middlebury. 

\section{Conclusion}
\label{sec:conclusion}

In this paper, we have presented a novel recurrent stereo-matching framework by integrating a foundation model, Depth Anything V2. We developed simple and effective techniques to harness its powerful pre-retrained feature representations, enhancing both the matching feature and context extraction in recurrent stereo matching. Despite its universal robustness, we still observed that when evaluating its depth estimate on standard stereo datasets, scale inconsistency across different image regions exists, particularly for the synthetic ones. To address this in utilizing the depth estimate to facilitate the recurrent disparity, we introduced a scale update module designed to iteratively recover dense scale maps. Our model has been verified to have significant effectiveness on both in-domain fitting and zero-shot generation and achieves leading performance in standard stereo benchmarks. We believe that the work presented in this paper contributes to advancing the exploration of the foundational model for stereo matching.

{\small
\bibliographystyle{ieeenat_fullname}
\bibliography{sec/11_references}
}

\ifarxiv \clearpage \maketitlesupplementary \appendix

\section{Evaluation of Depth Anything V2}
\label{sec:danv2}

In this section, we present the evaluation of Depth Anything V2~\cite{depth_anything_v2} on some common stereo datasets. 
To achieve cross-scene robustness, the relative depth estimation models are usually trained with a scale and shift (affine) invariant loss~\cite{ranftl2020towards} on the inverse depth space, thus predicting an affine disparity with unknown scale and shift.  
Given an inverse depth map predicted by Depth Anything V2 is $\mathbf{z}$ and its corresponding ground truth disparity map $\mathbf{d}_{gt}$, they must satisfy the following affine transformation,
\begin{equation}
\mathbf{d}_{gt} = s\mathbf{z}+t, 
\label{eq:af}
\end{equation}
where $s$ is the scale and $t$ is the shift.
For each image example, we can use the least square to find the solution for the scale and shift, $\hat{s}$ and $\hat{t}$. The aligned disparity map from the depth estimate is computed as,
\begin{equation}
\hat{\mathbf{d}}= \hat{s}\mathbf{z}+\hat{t}.
\label{eq:ad}
\end{equation}

The quality of the depth estimate of Depth Anything V2 can be accessed with the end-point error (EPE) between $\mathbf{d}_{gt}$ and $\hat{\mathbf{d}}$. 
To evaluate the scale consistency, we further compute a ratio map between the ground truth disparity $\mathbf{d}_{gt}$ and the aligned disparity map $\hat{\mathbf{d}}$, 
\begin{equation}
\mathbf{r}= \frac{\mathbf{d}_{gt}}{\hat{\mathbf{d}}}, 
\label{eq:ratio}
\end{equation}
where $\hat{\mathbf{d}}$ is clamped with a minimum (set to 0.05) in advance to avoid meaningless division. 

The distribution of the values in $\mathbf{r}$ reveals the scale consistency. If the depth estimate were scale-consistent, most values in $\mathbf{r}$ approximate to 1, otherwise there must be many ratios that deviate to 1. Therefore, we compute the standard deviation of  $\mathbf{r}$ to assess the scale consistency. 

We perform evaluation on three realistic datasets, KITTI 2015~\cite{menze2015object},  Middlebury (half resolution)~\cite{scharstein2014high} and ETH3D~\cite{schops2017multi}, and two synthetic datasets, Scene Flow~\cite{sceneflow} and CREStereo~\cite{li2022practical}. For the realistic datasets, we evaluate on their entire trainsets, \ie, 200 examples for KITTI 2015, 15 samples for Middlebury, and 27 samples for ETH3D. For Scene Flow and CREStereo, we evaluate on 200 random samples from their trainsets.

Tab.~\ref{tab:danv2} presents the quantitative results. Even given the unknown scale and shift, disparity errors are large, especially for the synthetic Scene Flow~\cite{sceneflow} and CREStereo~\cite{li2022practical} datasets, whose STD is very high too, indicating the scale inconsistency within the image is serious. This is because the synthetic stereo dataset is about unnatural scenes. 
As the stereo model is usually pre-trained on the synthetic stereo dataset, the scale inconsistency of DEFOM poses challenges for recovering disparity from its depth estimate. 
In contrast, Depth Anything V2 presents better results on realistic datasets. Although the EPE on Middlebury-half is up to 5, it is mainly due to its high resolution. In contrast, for two synthetic datasets, both the EPE and STD are larger, indicating Depth Anything V2 does not predict depth maps with good scale consistency.  

Fig.~\ref{fig:dav2} visualizes examples from Scene Flow, KITTI, Middlebury, and ETH3D. The last column is the ratio map with a color bar. These visualizations further highlight the scale inconsistency issue, especially in the Scene Flow dataset, though some scale inconsistency is also evident in the real datasets, albeit to a lesser extent. Despite this, synthetic datasets, like Scene Flow, help to train the scale update module, as they pose more challenges to scale recovery.

\begin{table}
  \centering
\resizebox{1.0\columnwidth}{!}{
  \begin{tabular}{cc|cc|cc|cc|cc}
    \toprule
    \multicolumn{2}{c|}{Sceneflow} & \multicolumn{2}{c|}{CREStereo} &\multicolumn{2}{c|}{KITTI-2015} &\multicolumn{2}{c|}{Middlebury-half} &\multicolumn{2}{c}{ETH3D}\\
    EPE & STD & EPE & STD & EPE & STD & EPE & STD & EPE & STD \\
    \midrule
    8.04 & 3.31 & 5.10 & 3.30 & 2.08 & 0.74 & 5.00 & 0.11 & 0.65 & 0.40 \\
    \bottomrule
  \end{tabular}
}
  \vspace{-5pt}
  \caption{Examination of Depth Anything V2 on typical stereo datasets via least-square affine alignment.}
  \label{tab:danv2}
  \vspace{-15pt}
\end{table}

\begin{figure*}[t]
\centering
\input{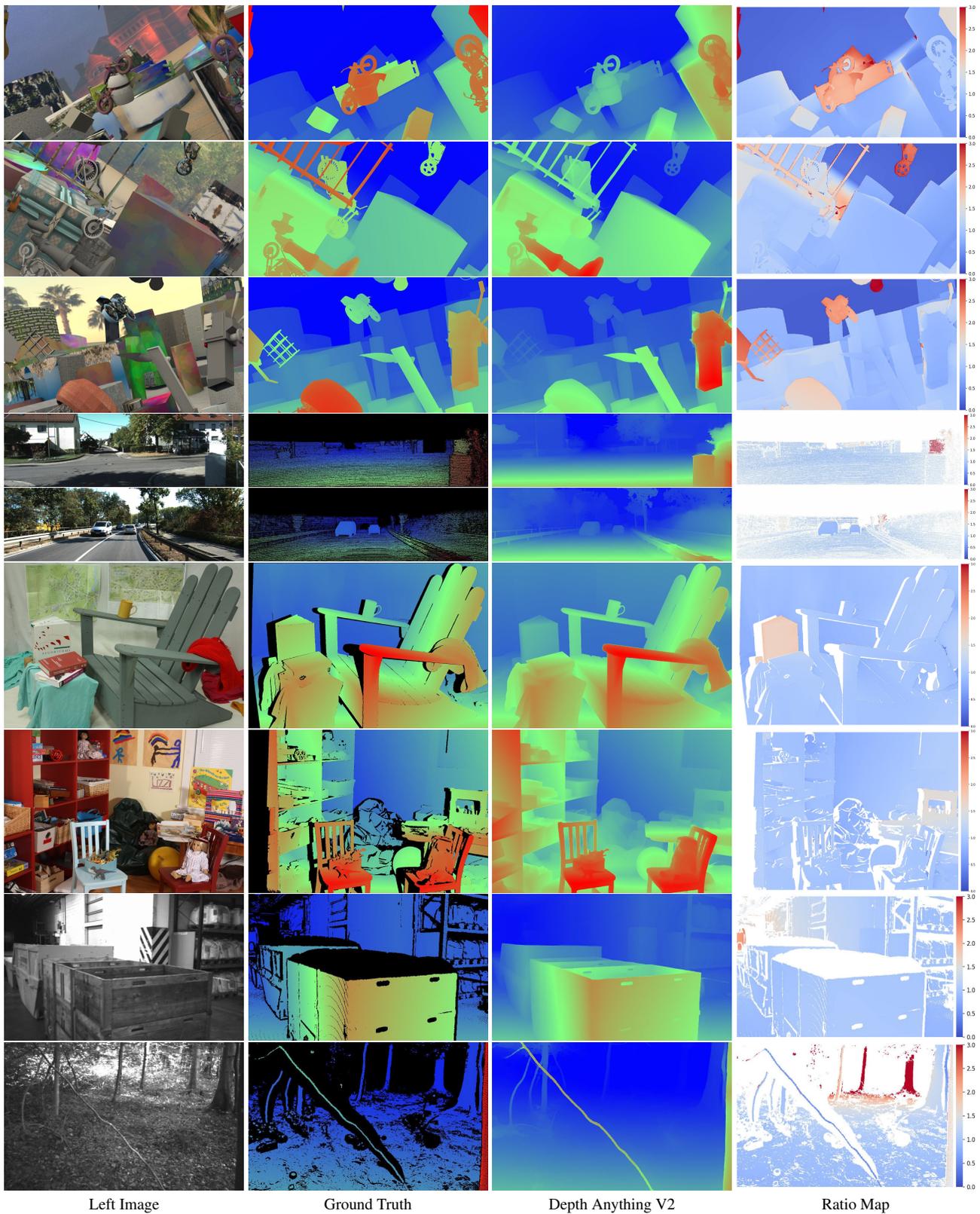}
\vspace{-5pt}
\caption{Visualization of the aligned depth estimate of Depth Anything V2 on some examples of the stereo datasets. \textbf{Row 1-3:} Scene Flow. \textbf{Row 4:} KITTI 2012. \textbf{Row 5:} KITTI 2015. \textbf{Row 6-7:} Middlebury. \textbf{Row 8-9:} ETH3D. Best viewed in color and by zooming in.}
\label{fig:dav2}
\end{figure*}

\newpage
\section{Additional Ablation Study}

\subsection{Combined Encoders}

In this section, we present an ablation study about combined encoders's design choices. Tab.~\ref{tab:coen} shows the results. For both the combined feature encoder and context encoder, we simultaneously experiment with other design choices for them, including using the DPT only to construct the encoders without CNNs and using the original fixed DPT instead of a new trainable DPT.

\begin{table}[!ht]
  \centering
\resizebox{1.0\columnwidth}{!}{
  \begin{tabular}{l|cc|c|c|c}
    \toprule
    \multirow{2}{*}{Models} & \multicolumn{2}{c|}{Scene Flow} & {KITTI 2015} & {Midd.-half} & {ETH3D} \\
     & EPE & Bad 1.0 & Bad 3.0 & Bad 2.0 & Bad 1.0 \\
    \midrule
      without CNNs  & 0.619 & 7.078 & 5.998 & 9.379 & 3.872  \\
      without new DPT  & 0.473 & 5.815 & 5.450 & 7.959 & 2.278 \\
      Full Model  & 0.458 & 5.571 & 5.289 & 6.760 & 2.614  \\
   \bottomrule
  \end{tabular}
  }
  \caption{\textbf{Ablation study on the design choices combined encoders.} Midd.-half represents Middlebury (half resolution).}
  \label{tab:coen}
\end{table}

\begin{figure*}[!ht]
\centering{
\input{figs/zeroshot_sup.tex}
}
\vspace{-18pt}
\caption{Zero-Shot qualitative comparison with RAFT-Stereo~\cite{lipson2021raft} Mocha-Stereo~\cite{chen2024mocha} and Selective-IGEV~\cite{wang2024selective} on the four common realistic stereo datasets.  \textbf{Row 1-2:} KITTI 2012. \textbf{Row 3-4:} KITTI 2015. \textbf{Row 5-6:} Middlebury-full. \textbf{Row 7-8:} Middlebury-half. \textbf{Row 9-10:} ETH3D. Best viewed in color and by zooming in.}
\label{fig:zeroshot_sup}
\vspace{-8pt}
\end{figure*}

\textbf{Can we simply abandon CNNs?}  The answer is \textbf{No}. We first ablate the CNNs in the encoders and use the feature maps from the new DPT head only as the matching feature maps and context maps. The results are listed in the first row of Tab.~\ref{tab:coen}. The ablation would result in a significant performance drop on both in-domain test and zero-shot generation. For example, The EPE increases by over $35\%$ on Scene Flow, and Bad 2.0 increases by over $35\%$ on Middlebury (half resolution). The results indicate that the CNN feature is still necessary for the proposed model. 

\textbf{Is a new DPT beneficial?} The answer is \textbf{overall Yes}. The second row of Tab.~\ref{tab:coen} shows the result of the model that used the fixed DPT of Depth Anything V2. There are about $3-5\%$ error increases on Scene Flow and KITTI 15 and a $23\%$ rise on Middlebury (half resolution), while a $13\%$ drop on ETH3D. We thus hypothesize that the fixed DPT is more favorable to data with a small disparity range ($<64$), like ETH3D and a new trainable DPT is more helpful to the large disparity. As a new trainable DPT is generally better, we include it in our final model.

\begin{table}[!ht]
\small
\centering
\resizebox{1.0\columnwidth}{!}{
\begin{tabular}{l| c c c c c c c c}
\toprule 
SU Iter & 0 & 1 & 3 & 5 & 7 & 8 & 9 & 10 \\
\midrule
{EPE}     & 0.752 & 0.668 & 0.651 & 0.650 & 0.640 & 0.636 & 0.637 & 0.660 \\
{Bad 1.0} & 9.018 & 8.030 & 7.804 & 7.709 & 7.683 & 7.697 & 7.660 & 8.243 \\
\bottomrule
\end{tabular}
}
\vspace{-0.5em}
\caption{\textbf{Ablation study on scale update iterations.}}
\vspace{-1.2em}
\label{tab:sui}
\end{table}

\subsection{Iterations of Scale Update}
In this section, we investigate the effect of the number of iterations of the scale update module in Tab.~\ref{tab:sui}. Likewise, we fixed the total number of iterations as 18 in training and 32 in evaluation, and the number of SU iterations is set the same in training and evaluation. We perform training on Scene Flow for 50k steps with a batch size of 4 and evaluation on the Scene Flow test set. The number of scale update iterations is increased from 0 to 10 and the 0 scale update iteration represents that the scale update module is not used. When the scale update iteration is 0, the model has the highest error metrics, and increasing it to 1 results in over $12\%$ error reduction. The EPE continues to decrease until the scale update iteration reaches 8 and the performance for the 7-9 scale update iteration is closed. When the scale update iteration exceeds 9, the performance starts to drop obviously. Therefore, we select 8 as the number of scale update iterations in our final model.


\section{Zero-Shot Qualitative Comparison}
In this section, we provide more visual comparison with RAFT-Stereo~\cite{lipson2021raft} , Mocha-Stereo~\cite{chen2024mocha} and Selective-IGEV~\cite{wang2024selective} Therefore, we provide a visual comparison among RAFT-Stereo~\cite{lipson2021raft} with our DEFOM-Stereo (VIT-S), and DEFOM-Stereo (VIT-L). Fig.~\ref{fig:zeroshot_sup} presents the comparison on four common stereo datasets. We also provide the qualitative comparison on a more diverse stereo image dataset Flickr1024~\cite{wang2019flickr1024} in Fig.~\ref{fig:flickr1024} and Fig.~\ref{fig:flickr1024B}. All the models are pre-trained on the Scene FLow dataset only. The clear advantages of our models can be seen in the visual comparison.

\begin{figure*}[t]
\centering{
\input{figs/flickr1024.tex}
}
\caption{Zero-Shot qualitative comparison with RAFT-Stereo~\cite{lipson2021raft}, Mocha-Stereo~\cite{chen2024mocha} and Selective-IGEV~\cite{wang2024selective} on Flickr1024~\cite{wang2019flickr1024}. Best viewed in color and by zooming in.}
\label{fig:flickr1024}
\end{figure*}

\begin{figure*}[t]
\centering{
\input{figs/flickr1024B.tex}
}
\caption{Zero-Shot qualitative comparison with RAFT-Stereo~\cite{lipson2021raft}, Mocha-Stereo~\cite{chen2024mocha} and Selective-IGEV~\cite{wang2024selective} on Flickr1024~\cite{wang2019flickr1024}. Best viewed in color and by zooming in.}
\label{fig:flickr1024B}
\end{figure*}

\section{Qualitative Comparison of RVC Models}
This section presents a visual comparison among robust vision challenge models. We compare our RVC model with the previous best-performing model of individual benchmarks, \ie, UCFNet\_RVC~\cite{shen2023digging} on KITTI 2015, CREStereo++\_RVC~\cite{jing2023uncertainty} on Middlebury and LoS\_RVC~\cite{li2024local} on ETH3D. And our model demonstrates more accurate results simultaneously. 

\begin{figure*}[t]
\centering{
\input{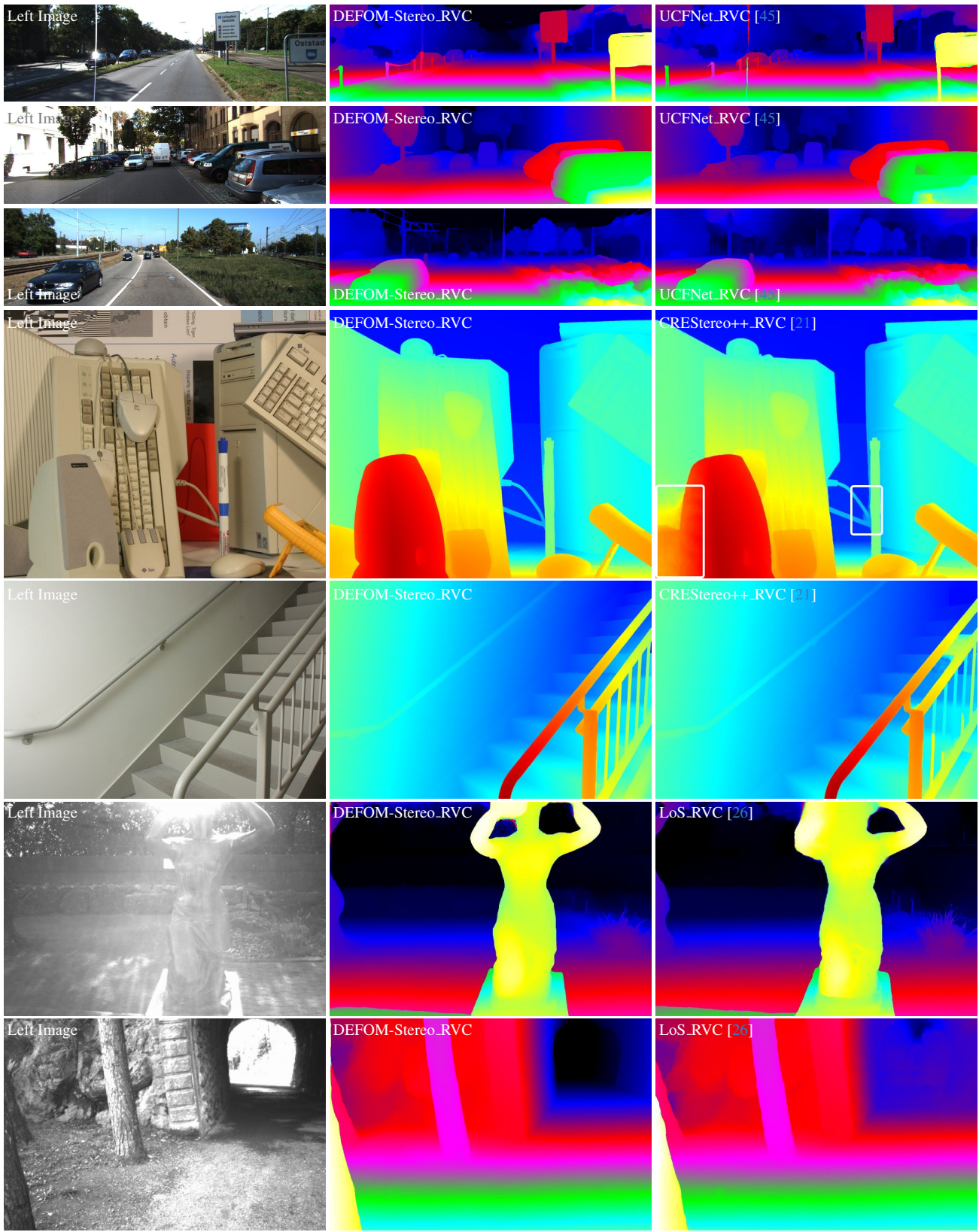}
}
\caption{Qualitative Comparison among top-performing RVC models, including UCFNet\_RVC~\cite{shen2023digging}, CREStereo++\_RVC~\cite{jing2023uncertainty}, LoS\_RVC~\cite{li2024local} and our model. \textbf{Row 1-3:} KITTI 2015. \textbf{Row 4-5:} Middlebury. \textbf{Row 5-6:} ETH3D. Best viewed in color and by zooming in.}
\label{fig:rvc}
\end{figure*}

\begin{table}[!ht]
  \centering
\resizebox{1.0\columnwidth}{!}{
  \begin{tabular}{l|c|c|c|c}
    \toprule
    Methods & All(100\%) & Non-occluded (87.92\%) & Occluded(12.08\%) & Textureless(59.60\%) \\
    \midrule
    Our Baseline & 13.44 & 10.64 & 30.33 & 13.01  \\
    Mocha-Stereo & 11.49 & 9.11 & 25.79 & 12.25  \\
    Ours (ViT-S) & 6.76 & 4.29 & 20.83 & 7.05  \\
    Ours (ViT-L) & 5.91 & 3.26 & 20.64 & 6.04  \\
    \bottomrule
  \end{tabular}
}
  \vspace{-5pt}
  \caption{\footnotesize Zero-shot evaluation (Bad 2.0) on different areas of Midd.-half.}
  \label{tab:illposed}
  \vspace{-10pt}
\end{table}

\section{Evaluation on Ill-Pose Regions}

\newcommand\rone{\textcolor{cyan}{R-6m4P}}
\newcommand\rtwo{\textcolor{magenta}{R-qv2k}}
\newcommand\rthr{\textcolor{green}{R-Tq9J}}
\definecolor{myyellow}{rgb}{1, 1, 0}

\begin{figure*}[t]
\centering{
\input{figs/illpose.tex}
}
\vspace{-10pt}
\caption{Visual comparison on ill-posed areas. \textbf{Odd Rows:} Left Image and Disparity Maps. \textbf{Even Rows:} Region Masks and Error Maps. Non-occluded and textured regions are in \textcolor{red}{red}. Non-occluded and textureless regions are in \textcolor{myyellow}{yellow}. Occluded and textured regions are in \textcolor{blue}{blue}. Occluded and textureless regions are in \textcolor{cyan}{cyan}. Best viewed in color and by zooming in.}
\label{fig:illpose}
\vspace{-5pt}
\end{figure*}

To further show the detailed improvement in occluded and textureless areas, we evaluate the models on Middlebury, as the indoor scene contains sufficient occluded and textureless regions. We follow LoS to use SSIM to extract textureless regions from the image. Tab.~\ref{tab:illposed} shows the results, where the proportions of different regions are also counted. There is obvious improvement in these ill-posed areas. Fig.~\ref{fig:illpose} visualizes some examples for the evaluation.

\section{Transparent or Mirror Surfaces}
We follow the reviewer's suggestion to evaluate our model on the dataset about transparent or mirror (ToM) surfaces. We examine the models on the Booster datasett~\cite{zamaramirez2022booster} which features reflective and glass surfaces, and some examples are shown in Fig.~\ref{fig:limitations}. We find that DAv2 performs usually well for these reflective and transparent materials and our model also works if these ill-posed factors are not too serious. When there is a large mirror, our model cannot work, and DAv2(ViT-S) also slightly fails. For readers who require a very robot model for ToM, we refer them to Stereo Anywhere~\cite{bartolomei2024stereo}, which is specifically designed for this problem. 

\begin{figure*}
    \centering{
    \input{figs/tom.tex}}
    \vspace{-9pt}
    \caption{Examination of the depth models on the Booster dataset. Best viewed in color and by zooming in. }
    \label{fig:limitations}
\vspace{-9pt}
\end{figure*}

 \fi

\end{document}